\pdfoutput=1

\documentclass[11pt]{article}

\usepackage{acl} 
\usepackage{tikz}
\usepackage{pgfplotstable}
\usepackage{pgf-pie}
\usepackage{times}
\usepackage{latexsym}

\usepackage[T1]{fontenc}

\usepackage[utf8]{inputenc}

\usepackage{microtype}

\title{Towards End-User Development for IoT: A Case Study on Semantic Parsing of Cooking Recipes for Programming Kitchen Devices}

\author{Filippos Ventirozos\textnormal{,} Sarah Clinch \and Riza Batista-Navarro \\
        Department of Computer Science \\ School of Engineering \\ University of Manchester, United Kingdom\\
         \{filippos.ventirozos,sarah.clinch,riza.batista\}@manchester.ac.uk}

\usepackage[absolute]{textpos}

\textblockorigin{0mm}{0mm}

\begin{document}
\maketitle
\begin{abstract}
Semantic parsing of user-generated instructional text, in the way of enabling end-users to program the Internet of Things (IoT), is an underexplored area. 
In this study, we provide a unique annotated corpus which aims to support the transformation of cooking recipe instructions to machine-understandable commands for IoT devices in the kitchen.
Each of these commands is a tuple capturing the semantics of an instruction involving a kitchen device in terms of ``What'', ``Where'', ``Why'' and ``How''.
Based on this corpus, we developed machine learning-based sequence labelling methods, namely conditional random fields (CRF) and a neural network model, in order to parse recipe instructions and extract our tuples of interest from them. 
Our results show that while it is feasible to train semantic parsers based on our annotations, most natural-language instructions are incomplete, and thus transforming them into formal meaning representation, is not straightforward.
\end{abstract}

\section{Introduction}

The Internet of Things (IoT) has emerged as a powerful ecosystem for supporting people in their day-to-day activities. 
IoT refers to ``internet-connected devices which can see, hear, think and perform jobs by having them talk with each other, to share information and to synchronise pronouncements'' \citep{shah_survey:_2016}.
Some examples of those \textit{devices} are sensors and actuators, mobile phones, tablets and smart home appliances.

With the proliferation of IoT devices, programming them to perform user-defined jobs (or tasks) has become a challenge.
Given that the users of such devices would have varying requirements and preferences, employing professionals to write programs for controlling them will, in most cases, not be a viable, cost-effective option \citep{PUCE}.
It is practically impossible for professional designers and developers to keep up with the expectations of end-users, especially considering the wide range of possible applications and contexts of use \citep{EUP_new_IoT}.

To eliminate the aforementioned bottleneck, studies in end-user development (EUD) have emerged, aimed at enabling end-users to ``program'' IoT devices, without requiring them to learn any programming languages \citep{EUP_PIP,Paterno2017_IoT}.
One EUD approach is known as meta-design, whereby systems which were not designed a priori, are programmed based on the participation of end-users \citep{meta_is_the_future}.
In a related method called programming by demonstration or programming by example, a machine carries out particular tasks by learning from user-provided demonstration (or instructions), rather than by receiving machine commands \citep{Paterno2017_IoT}.

In this work, we investigate the feasibility of using programming by demonstration to enable end-users to program IoT devices, on the basis of written (textual) instructions.
Specifically, we focus on the case of IoT devices in the kitchen, and have developed a method for semantic parsing of cooking recipes and converting them into IoT commands, to enable machines to carry out kitchen-related tasks.

\begin{textblock}{20}(7,28) 
\noindent \Large{*Work was completed in January 2020*}
\end{textblock}

Compared to other areas in the household, the kitchen is believed to have been left behind in terms of automation \citep{MAMPF}. 
A study by \citet{coskun_is_2018} found that end-users would like to effortlessly interact with kitchen appliances, as this would enhance their cooking experience and make their food preparation tasks more efficient and flow more seamlessly.
However, currently available \textit{smart} (i.e. IoT-enabled) kitchen devices \citep{kognichef,schneider_semantic_2007,MAMPF} require users to ``program'' them by specifying values for various settings manually.
We seek to eliminate this manual effort by enabling devices to learn this knowledge from cooking recipes, using a semantic parsing method.  
To the best of our knowledge, this work constitutes the first approach underpinned by natural language processing (NLP) methods, that is aimed at reducing the effort from end-users in programming IoT devices in the kitchen. 

The contributions of the work reported in this paper include:
(a) a first-of-its-kind corpus, in which natural-language recipe instructions have been labelled with semantics derived from IoT semantic control theory; (b) machine learning approaches for semantic parsing of cooking recipes; and (c) an analysis of the ``completeness'' of recipe instructions, answering the question of whether recipes contain all of the necessary information for generating fully formed, machine-understandable IoT commands.

\section{Related Work}

Previous studies have investigated the feasibility of enabling end-users to program IoT devices through gestures and spoken instructions \citep{chen17improv,mofrad18speech}.
However, the use of natural-language text as the basis of end-user programming has been underexplored.
Frameworks such as InterPlay \citep{messer06interplay} and AppsGate \citep{coutaz16first} were developed to allow end-users to control devices in a smart home, but using structured (template-based) instructions, rather than natural language.
\citet{perera15natural} collected natural-language household-related reminders from end-users, in the form of sticky notes. They then studied their linguistic features but did not attempt to automatically transform them into any formal, machine-understandable instructions.

In this work, we are focusing on semantic parsing of cooking recipes, a specific type of instructional text, i.e., natural-language descriptions of steps for performing specific tasks. Semantic parsing---the transformation of unstructured, natural language into a structured, formal meaning representation \citep{andreas-etal-2013-semantic}---lends itself well to the task of generating machine-understandable IoT commands from texts.
For instance, it has been applied to the conversion of natural-language text queries to machine-understandable code such as regular expressions \citep{kushman-barzilay-2013-using} and queries in the Structured Query Language (SQL) format \citep{yu-etal-2018-spider}.

Instructional text contains procedural knowledge, which is crucial in facilitating task automation by machines.
However, instructional texts tend to be incomplete in that they leave it to the reader to infer missing script knowledge, i.e., standard event sequences that comprise typical everyday activities \citep{wanzare-etal-2016-crowdsourced}. For this reason, much of previous work analysing cooking recipes has focussed on understanding temporal, causal and dependency relationships between events, i.e., food preparation steps \citep{abend-etal-2015-lexical,yordanova-2015-discovering,jermsurawong-habash-2015-predicting}. Our work, however, aims to analyse cooking recipes at a finer-grained level. That is, we seek to perform semantic parsing on each recipe instruction (involving a kitchen device), to convert it into a formalised, machine-understandable representation.

Most similar to our research is the work of \citet{zhang-etal-2012-automatically}.
In their proposed approach, handcrafted rules were developed to identify instruction components (e.g., instrument, purpose, quantitative and spatial parameters) based on the outputs of the Stanford dependency parser.
In contrast, our work builds upon machine learning-based methods, described in Section~\ref{chapter:semantic_parsing}, for recognising instruction components.
Furthermore, our target formalisation (i.e., the representation into which natural language is to be transformed) is grounded on a convention that was found to be suitable for capturing IoT semantics \citep{semiotics}.

\section{Methodology}
\label{chapter:methodology}
In this section, we present the methods underpinning our proposed approach, starting with our chosen formal representation, followed by data collection, the annotation task and the development of semantic parsers. 

\subsection{Formal Representation}
\label{chapter:formal_representation}
Choosing a target formal representation that captures IoT semantics, into which natural-language instructions will be transformed, is a crucial first step. We adopted the 5W1H representation \citep{4w1h_method_1,4w1h_method_2}, which can be used to describe any event or situation in terms of answers to the questions ``What'', ``Where'', ``When'', ``Who'' and ``How'' (hence its name). We draw inspiration from a previous study which demonstrated the suitability of 3W1H for representing concepts related to the control of IoT devices  \citep{semiotics}, adapting it to the kitchen environment in particular. 
To this end, we established the following definitions, with examples based on the sample natural-language instruction \emph{``Increase the temperature of the oven to 400 degrees Fahrenheit''}:
\begin{itemize}
    \item Where: the specific kitchen device where the instruction should be carried out, e.g., \emph{``oven''}
    \item What: a property of the device of interest, e.g., \emph{``temperature''}
    \item Why: the purpose or role, e.g., \emph{``Increase''}
    \item How: the value of the property and its unit of measurement, e.g., \emph{``400 Fahrenheit''}
\end{itemize}

We note that for our purposes, there are only three out of the traditional five Ws. The answer to the question ``Who'' remains consistent — it always refers to the reader of the instruction who is expected to execute it. Additionally, the ``When'' is deemed outside the purview of our tuple as per \citep{semiotics}. Any temporal references encountered are labelled as ``How''. For example, in the statement \emph{``bake for \underline{40 minutes}''}, the duration \emph{``40 minutes''} is attributed to the ``How'' component of the tuple.


We posit that the above representation is sufficient for capturing the semantics of any given recipe instruction, and thus develop an annotation task based on it, described in more detail in Section~\ref{sec:annotation} below.

\subsection{Data Collection and Pre-processing}
For our work, we sought to construct an annotated corpus of recipes based on a subset of the publicly available RecipeQA dataset \citep{yagcioglu-etal-2018-recipeqa}, which contains around 20,000 unique cooking recipes in English. As a first step, we extracted titles and instructions from the recipes in the dataset (leaving out any images and the lines of text corresponding to question answering).
The following pre-processing steps were then performed:
(a) conversion from UTF-8 to ASCII character encoding, to eliminate any noise from special characters; (b) conversion of acronyms that are typically used in cooking to their respective full forms (e.g., from \emph{``w/o''} to \emph{``without''}) using a lookup list; and (c) removal of emoticons, extraneous punctuation marks and consecutive spaces.

Out of the 20,000 recipes, we chose only a subset mentioning kitchen devices. 
To this end, we compiled a dictionary of words related to kitchen devices to guide our selection.
Specifically, we searched for relevant vocabularies and identified the DogOnt ontology \citep{dogont} as the only ontology that catalogues home automation devices (including IoT devices in the kitchen).
Additionally, we retrieved all of the hyponyms of \emph{``kitchen appliance''} in WordNet \citep{10.1145/219717.219748}, and mapped them to device classes in DogOnt, in order to organise them hierarchically. 
The result is a total of 10 upper-level kitchen device classes, with \emph{``oven''} and \emph{``stove''} having the highest number of subclasses (six and eight, respectively).
To further enrich our dictionary, we made use of GloVe word embedding vectors \citep{pennington-etal-2014-glove} pre-trained on Common Crawl\footnote{\href{https://commoncrawl.org}{commoncrawl.org}}, to obtain 100 words that are most similar to each of the 10 higher-level device classes.
These words were manually reviewed and those which are not related (by synonymity or device purpose) were removed. 
In the work described in this paper, we focussed on only two device classes, namely, \emph{``oven''} and \emph{``fridge''}\footnote{The dictionaries for these two device classes can be found at \href{https://github.com/FilipposVentirozos/Towards-End-User-Development-for-IoT}{github.com/FilipposVentirozos/Towards-End-User-Development-for-IoT}.}.

\subsection{Annotation Task}
\label{sec:annotation}
Our annotation scheme was derived from the work of \citet{semiotics}, as explained in Section \ref{chapter:formal_representation}. 
For every recipe instruction, any information pertaining to location was labelled as ``Where''; in our case, this is a mention of a kitchen device.
In the same sentence, the sequence of words corresponding to each of ``What'', ``Why'' and ``How'' was also annotated. 
This task, which we cast as a sequence labelling task, was facilitated by the \textit{doccano} annotation tool \citep{doccano}. It comes with a user-friendly interface for sequence labelling, allowing annotators to select any word sequence and to label it as one of ``Where'', ``What'', ``Why'' and ``How''.

\subsubsection{Annotation Guidelines}

\label{sec:annotation-guidelines}
We established the annotation guidelines before any of the annotators were asked to carry out the tasks. The key points are outlined below.

\begin{itemize}
    \item The value of ``Where'' should be in our compiled dictionary.
    \item  Only instructions directly related to a kitchen device are of interest. For instance, in the instruction ``\textit{After 10 minutes, take out the dish from the oven, stir, and then set it back inside}'', the instructions of interest are those pertaining to setting the oven timer and the (implicit) opening of the oven, rather than the stirring of the dish.
   \item Only instructions where the device is used to cook or produce the dish are annotated. For example, in the instruction ``\textit{Take out eggs from the fridge}'', the fridge was used for storage and not as a device for producing the dish; hence, this instruction is not annotated. This guideline would also exclude comments such as ``\textit{my microwave is 1000 Watt}''.

\end{itemize}

The annotations were carried out by three annotators.
One annotator (the first author of this paper) completed the sequence labelling task for 221 recipes, while the second and third annotators completed 65 and 49 recipes, respectively.
A common subset of 10 recipes (containing 67 instructions) were annotated by all of them, with an inter-annotator agreement (IAA) of 59\%  in terms of F1-score.

\subsection{Semantic Parsing}
\label{chapter:semantic_parsing}

Below, we describe in detail our machine learning-based approaches to semantic parsing. Specifically, the sequence labelling task for recognising the values of ``Where'', ``What'', ``Why'' and ``How''.

\subsubsection{Conditional Random Fields}
\label{chapter:crf}
Firstly, we developed sequence labelling models based on conditional random fields (CRFs) \citep{CRFsuite}.
Recipe instructions were tokenised, lemmatised and tagged with parts of speech (POS) using the spaCy library \citep{spacy2}.
The characters of each token were then converted to lowercase.
Afterwards, the following features were extracted for each token: the token itself, its lemma, its first letter, its first three letters and its  POS tag. Additionally, we also made use of binary features: whether a token is capitalised, exists in our compiled dictionary, consists of alphabetic characters, and whether it is a stop word.
Furthermore, we explored extracting the above features for a token's surrounding words (within a window of three), as well as its head token (as provided by the spaCy syntactic dependency parser).

\subsubsection{Neural Network}
We sought to determine how a neural network-based approach would perform in our sequence labelling task. 
To accomplish this we employed an open-source named entity recognition tool called NeuroNER \citep{dernoncourt-etal-2017-neuroner}. This tool is underpinned by a type of recurrent neural network (RNN) known as long short-term memory (LSTM) \citep{neuroner_architecture}. 
We configured it to make use of spaCy for tokenisation (same as in our CRF-based method described above), as well as pre-trained GloVe embeddings \citep{pennington-etal-2014-glove} for learning word sequence representations. We trained a new NeuroNER model on our own annotated corpus, using the default hyper-parameters for the neural network\footnote{\href{https://github.com/Franck-Dernoncourt/NeuroNER/blob/master/parameters.ini}{github.com/Franck-Dernoncourt/NeuroNER}}.

\section{Evaluation and Results}

\subsection{Evaluation Method}
\label{section:eval_method}
We collected all of the recipes about a given device (according to the manual annotations) and tokenised them using the spaCy tool \citep{spacy2}. 
These recipes were then concatenated, resulting in a total of 195 and 120 recipes mentioning the oven and fridge, respectively.
Each word is labelled according to the IOB scheme \citep{ramshaw-marcus-1995-text}, with four semantic labels: ``Where", ``What", ``Why" and ``How".
In total there are nine possible labels: ``B-Where'', ``I-Where'', ``B-What'', ``I-What'', ``B-Why'', ``I-Why'', ``B-How'', ``I-How'' and ``O''.

We partitioned the dataset into training, validation, and test subsets using splits of 70\%, 15\%, and 15\% respectively\footnote{The datasets can be found at \href{https://github.com/FilipposVentirozos/Towards-End-User-Development-for-IoT}{github.com/FilipposVentirozos/Towards-End-User-Development-for-IoT}.}. To guarantee a representative distribution in each subset, stratification was conducted based on the nine designated labels. Additionally, this stratification process ensured an even distribution of false positive recipes. False positive recipes refer to those containing a "Where" term from the dictionary, yet upon annotation, were not perceived to contain instructions related to the targeted devices.


Our proposed methodologies were subjected to evaluation using the test set. The validation set was used in the following manner: (1) to determine the epoch for stopping the training of the neural network; and (2) to select optimal features and tune hyper-parameters for CRFs. For the latter, we measured F1-score \citep{F1} by subtracting groups of features (see section \ref{chapter:crf}) in the following order: (1) parent set of features, (2) features of up to three surrounding words, (3) features of up to two surrounding words, and (4) features of only one surrounding word. In addition, we performed three-fold cross-validation on the training and validation set combined to find the best hyper-parameters for the CRF. For that a random search algorithm \citep{scikit-learn} was used with 80 possible combinations of the hyper-parameters $c1$, $c2$ and $min\_freq$, where the first two stand for L1 and L2 regularisation, respectively, and the last one a cut-off threshold for occurrence frequency of a feature\footnote{See \href{https://sklearn-crfsuite.readthedocs.io/en/latest/api.html}{sklearn-crfsuite.readthedocs.io/en/latest/api.html} for further details}.

\subsection{Results}
\subsubsection{Exploratory Analysis}
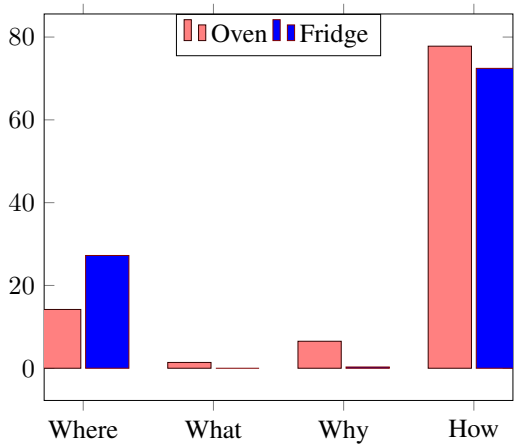
\begin{figure}
\centering
\pgfplotstableread[row sep=\\,col sep=&]{
    label & freq_oven & freq_fridge \\
    Where & 14.2 & 27.24  \\
    What  & 1.4  & 0  \\
    Why   & 6.52  & 0.3 \\
    How   & 77.8  & 72.44 \\
    }\mydata
\begin{tikzpicture}[scale=0.9] 
    \begin{axis}[
            ybar,
            bar width=18,
            symbolic x coords={Where,What,Why,How},
            xtick=data,
            legend style={at={(0.5,1)},
                anchor=north,legend columns=-1},
        ]
        \addplot [red!20!black,fill=red!50!white]  table[x=label,y=freq_oven]{\mydata};
        \addplot [red!50!black,fill=blue!100!white] 
        table[x=label,y=freq_fridge]{\mydata};
        \legend{Oven, Fridge}
    \end{axis}
\end{tikzpicture}
\caption{Distribution of the values for the four slots (``Where'', ``What'', ``Why'' and ``How''). The pink and blue bars correspond to the annotations for the two devices, oven and fridge, respectively.}
\label{fig:bar_slots}
\end{figure}

Figure \ref{fig:bar_slots} provides a depiction of the per-device distribution of word sequences labelled as ``Where'', ``What'', ``Why'' and ``How''. There were a total of 2,131 oven-related and 323 fridge-related labels.




\subsubsection{Prediction}
\label{sec:prediction}
In Tables \ref{table:oven_results} and \ref{table:fridge_results} are the predictions against the test set for the oven and the fridge, respectively. 
For the CRF approach, the best F1-score was achieved when all of the group of features were used; see Section \ref{section:eval_method}. 
It was noticeable, however, that adding more features caused the model score to plateau; hence, the predefined group of features were considered enough. 

\begin{table}
\begin{tabular}{llll}
\multicolumn{4}{c}{Oven}                                                                                                                        \\ \hline
\multicolumn{1}{|c|}{\begin{tabular}[c]{@{}c@{}}Label \&\\  Avg\end{tabular}} & \multicolumn{1}{c|}{\begin{tabular}[c]{@{}c@{}}CRF Suite\\ F1 Score\end{tabular}} & \multicolumn{1}{c|}{\begin{tabular}[c]{@{}c@{}}NeuroNER\\  F1 Score\end{tabular}} & \multicolumn{1}{c|}{\begin{tabular}[c]{@{}c@{}}N. \\samples\end{tabular}} \\ \hline
\multicolumn{1}{|l|}{How}                                                     & 0.550                                                                             & 0.573                                                                             & 226                          \\ \cline{1-1}
\multicolumn{1}{|l|}{What}                                                    & 0.000                                                                             & 0.000                                                                             & 3                            \\ \cline{1-1}
\multicolumn{1}{|l|}{Where}                                                   & 0.704                                                                             & 0.778                                                                             & 34                           \\ \cline{1-1}
\multicolumn{1}{|l|}{Why}                                                     & 0.944                                                                             & 0.878                                                                             & 19                           \\ \hline
\multicolumn{1}{|l|}{micro avg}                                               & \multicolumn{1}{l|}{0.598}                                                        & \multicolumn{1}{l|}{\textbf{0.624}}                                               & \multicolumn{1}{l|}{282}      \\ \hline

\end{tabular}
\caption{
Comparative results for conditional random fields and the neural network approach on the test set, with respect to the oven. The best performance is shown in bold. The last column displays the number of true labelled words in the test set.}
\label{table:oven_results}
\end{table}

\begin{table}[ht]
\begin{tabular}{llll}
\multicolumn{4}{c}{Fridge}                                                                                                                                                                                                                                                           \\ \hline
\multicolumn{1}{|c|}{\begin{tabular}[c]{@{}c@{}}Label \&\\  Avg\end{tabular}} & \multicolumn{1}{c|}{\begin{tabular}[c]{@{}c@{}}CRF Suite\\ F1 Score\end{tabular}} & \multicolumn{1}{c|}{\begin{tabular}[c]{@{}c@{}}NeuroNER\\  F1 Score\end{tabular}} & \multicolumn{1}{c|}{\begin{tabular}[c]{@{}c@{}}N. gold-\\ standard\\ samples\end{tabular}} \\ \hline
\multicolumn{1}{|l|}{How}                                                     & 0.488                                                                             & 0.760                                                                             & 29                          \\ \cline{1-1}
\multicolumn{1}{|l|}{Where}                                                   & 0.538                                                                             & 0.727                                                                             & 11                           \\ \cline{1-1}
\multicolumn{1}{|l|}{Why}                                                     & 0.00                                                                              & 0.000                                                                             & 0                           \\ \hline
\multicolumn{1}{|l|}{micro avg}                                               & \multicolumn{1}{l|}{0.507}                                                        & \multicolumn{1}{l|}{\textbf{0.750}}                                               & \multicolumn{1}{l|}{40}      \\ \hline

\end{tabular}
\caption{Comparative results for conditional random fields and the neural network approach on the test set, with respect to the fridge. The best performance is shown in bold. The last column displays the number of true labelled words in the test set.}
\label{table:fridge_results}
\end{table}

\section{Discussion}
\label{chapter:discussion}
In Figure \ref{fig:bar_slots}, one can notice the discrepancy in frequency between the ``How'' label and the rest for each device. It is worth noting at this point that part of this discrepancy is because the ``How'' element often was labelled as a whole phrase, whereas the rest tended to have only one word labelled. Taking this adjustment into consideration, the labels ``How'' and ``Where'' seem to be the key identifier for an event. However, according to \newcite{semiotics} we need all four of them to describe an event. Based on our observation, recipes use common sense to define events; for instance, if we take the phrase \emph{``preheat the oven on 200 degrees C''}, we ourselves would understand that the reason to do this is to increase the temperature, from where we can infer that to increase is the ``Why'' and the object that we want to increase (i.e. the temperature) is the ``What''. Hence, there needs to be a mapping of ``Where'' and ``How'' to the quadruple.

During annotation, it was discovered that there are a number of code statements which link the events with each other. For instance, \emph{``cook until golden brown''} would refer to a ``while'' condition, \emph{``insert into the microwave and heat with 10 second intervals until melted''} refers also to a ``while'' loop until a condition is met, and \emph{``if you have a 1000W, cook for 30 seconds, unless with a 800W you should let it for 40 seconds''} is a clear ``if'' statement. The keywords which indicate code statements such as in the above examples ``until'', ``intervals'' and ``if'' were not labelled. It was noted that often these words link together the events that could relate to IoT commands and functionality.

The linkage of cooking events and device events could also help to infer device settings from imprecise instructions. During the annotation process it came to our attention that many recipes were not written using precise language. A workaround to this would be cross-referencing amongst other recipes. That is, mentions of devices could be linked to cooking events and through these, similar device instructions could be inferred. 

Finally, it was shown in our evaluation of semantic parsing (Section \ref{sec:prediction}) that the best performing approach for each device was the neural network-based approach. We believe that this is due to the CRF model not being able to capture long dependencies between words. Indeed, in using the CRF model, when more than three neighbouring words were inserted, the performance (in terms of F1-score) plateaued.

\section{Conclusion}
We created a first-of-its-kind annotated corpus that links instructional text to IoT devices. The annotations in the corpus were inspired by the 3W1H methodology that has been used in previous studies to describe IoT events. We also demonstrated methods for semantic parsing and compared them. We concluded that the neural network approach was superior. We also observed that many recipe instructions are written in imprecise language, leading to missing information that can be potentially inferred.

\section*{Acknowledgements}
We would like to express our gratitude to ARM Ltd and the UK Engineering and Physical Sciences Research Council (EPSRC), Grant No. EP/S513842/1 (Studentship 2109081), for their generous financial support, which was instrumental in facilitating this research study.

\bibliography{anthology,acl2020}
\bibliographystyle{acl_natbib}

\end{document}